# An Independent Evaluation of Subspace Face Recognition Algorithms

Dhiresh R. Surajpal and Tshilidzi Marwala

*Abstract*— **This paper explores a comparative study of both the linear and kernel implementations of three of the most popular Appearance-based Face Recognition projection classes, these being the methodologies of Principal Component Analysis (PCA), Linear Discriminant Analysis (LDA) and Independent Component Analysis (ICA). The experimental procedure provides a platform of equal working conditions and examines the ten algorithms in the categories of expression, illumination, occlusion and temporal delay. The results are then evaluated based on a sequential combination of assessment tools that facilitate both intuitive and statistical decisiveness among the intra and inter-class comparisons. The 'best' categorical algorithms are then incorporated into a hybrid methodology, where the advantageous effects of fusion strategies are considered**.

## I. INTRODUCTION

A human face is an extremely complex, dynamic and deformable object with features that can vary considerably and rapidly over time. Skin coverage offers a non-uniform material that is often difficult to model [1] and that can change in response to the effects of emotion, temperature, reflectance properties and perspiration levels, thus creating a large variety and variability within the configurations of facial expression. Another avenue includes time-varying changes by measure of growth, facial hair, effects on the skin due to aging and skin colour changes attributed to ultraviolet exposure. A further complexity is introduced by artefact related changes such as change due to injury and fashion-related issues such as cosmetics, jewellery and hairstyles [1]. Appearance-based analysis, which is one of the oldest approaches, is still said to give the most promising results [2]. Among the most popular publicly available subspace approaches are the classes of Principal Component Analysis (PCA), Independent Component Analysis (ICA) and Linear Discriminant Analysis (LDA).

When looking at the performance of all these algorithms, it is interesting to note the often 'contradictory' and confusing claims that have been made in the literature. Bartlett *et al.* [3] and Liu *et al.* [4], for example, claim that ICA outperforms PCA, while Baek *et al.* [5] claim that PCA is better. Moghaddam [6] states that there is no significant difference. Beveridge et al. [7] claim that in their tests LDA performed uniformly worse than PCA, Martinez [8] states that LDA is better for some tasks, and Belhumeur *et al.* [9] claim that LDA outperforms PCA. While all these claims may in fact hold a good degree of truth, one should bear in mind that there were differing control factors surrounding each conclusion i.e. the actual task statement, the subspace distance metrics, dimensionality retention and the non-standardized database choices etc [10]. All these conclusions have led to much debate and confusion over the years, particularly for an individual who is new to the field of FR and subspace methodologies and who seeks a good comparative understanding of the available techniques. Very rarely are all the classes compared in the same investigation and almost never are all of their implementations considered. This paper serves to provide a platform of equal conditions upon which the popular appearance-based subspace techniques can be fairly and properly benchmarked

This investigation will compare the appearance-based methodologies of PCA, LDA and ICA in both linear and kernel projections. Also both ICA architectures I and II [11] as well as both the InfoMax and FastICA implementations will be reviewed. The four popular and most widely used distance measures of L1 (City Block), L2 (Euclidean), Cosine and Mahalanobis have been chosen as the comparative classification metrics. The performance effects of illumination, expression, occlusion and time variations will be compared across all the techniques to conclusively yield the 'best' algorithm in each category.

While it may also be true that a robust classifier could be designed to effectively handle any one of the performance influencing factors, it is extremely difficult for an appearance-based technique to robustly deal with all the influencing variations [12]. Each individual classifier has a different sensitivity to different changes in facial variation and as was reported by Phillips *et al* [11], appearance-based methods show different levels of performance for different subsets of images. In their analysis of ICA and PCA, Bartlett *et al* [3] also reported that when incorrect classifications were made, it was very rare that the algorithms assigned the same incorrect identity class. The above findings strongly suggest that different classifiers contribute different and hence complementary information about the classification task. A *fusion* scheme involving the different face classifiers, which integrates multiple sources of evidence is therefore more likely to yield an overall improvement in both the efficiency and accuracy of the identification system. And while this may not solve the problem regarding influencing factors, it will definitely alleviate the impact they have on performance levels. It is for this reason that this investigation will also propose a hybrid formulation that combines the 'best' algorithm from each category.

## II. BACKGROUND

A two dimensional image, *Γ(x,y)* of size *m* (rows) by *n* (columns) pixels can generally be represented by concatenating the raster ordered values to create a vector in an *N* dimensional *image space* ($R^{N=m \times n}$). This image space, however, constitutes a rather high-dimensional space and recognition therein would be deemed computationally

D. R. Surajpal is with the School of Electrical and Information Engineering at the University of Witwatersrand, Johannesburg, South Africa; e-mail: d.surajpal@ee.wits.ac.za).

Prof. T. Marwala is Carl and Emily Fuchs Chair of Systems and Control at the School of Electrical and Information Engineering at the University of Witwatersrand, Johannesburg, South Africa; e-mail: t.marwala@ee.wits.ac.za ).

infeasible. If, however, an image of an object (say a face) is considered to be a point in the image space, then a set of *M* facial images can be represented as a set of points (samples of probability distribution) in the same confined *subspace* [13]. Theoretically it is common to model this subspace as a lower-dimensional *principle manifold*, embedded in a higher dimensional image space, wherein the *intrinsic* dimensionality is determined by the number of degrees of freedom within the face space. The goal of subspace analysis is thus to extract the *principle modes* of the underlying manifold, while retaining as much information (energy) from the original images as possible [2]. By doing this, subspace methodologies ensure that computational efficiency and hence the successful viability of face recognition algorithms can be achieved.

### A. Principal Component Analysis (PCA)

Given an s-dimensional vector representation of each face in a training set of M images, PCA tends to find a t-dimensional subspace whose basis vectors correspond to the maximum variance direction in the original image space. This new subspace is normally lower dimensional (t << s). All images of known faces are projected onto the face space to find sets of weights that describe the contribution of each vector. To identify an unknown image, that image is projected onto the face space as well to obtain its set of weights. By comparing a set of weights for the unknown face to sets of weights of known faces, the face can be identified.

### B. Independent Component Analysis and Linear Discriminant Analysis

PCA considered image elements as random variables with Gaussian distribution and minimized second-order statistics. Clearly, for any non-Gaussian distribution, largest variances would not correspond to PCA basis vectors. Independent Component Analysis (ICA) [3] minimizes both second-order and higher-order dependencies in the input data and attempts to find the basis along which the data (when projected onto them) are – *statistically independent*. Bartlett et al. [3] provided two architectures of ICA for face recognition task: *Architecture I* (ICA1) - statistically independent basis images and *Architecture II* (ICA2) - factorial code representation.

Linear Discriminant Analysis (LDA) [9] finds the vectors in the underlying space that best discriminate among classes.

### C. Kernel Methods

Variations in face images due to viewpoint, illumination and expression changes have been proven to be highly complex and nonlinear in nature [5] and it has been observed that variations between face images of the same person due to illumination and pose are almost always greater than image variations between the different persons [14]. From a classification viewpoint linear approaches, which only describe information based on second order statistics [15], are therefore said to be suboptimal in terms of accurate data representation. Complete pattern variation is said to be captured within the non-linear relations between neighbouring (three or more) pixels [15]. These relationships are represented in terms of higher order statistics that are crucial in fully representing complex patterns [16].

First introduced by Aizerman *et al* [17], the 'kernel trick' was used to map the input space to, by means of a nonlinear function expressed as dot products, to a convenient feature space (Hilbert space) in which the input data is nonlinearly related [15]. It was not until recently that Schölkopf *et al* [18] extended the classical PCA algorithm to Kernel Principal Component Analysis (KPCA) that was shown to be able to extract non-linear features and in doing so provide better recognition results than PCA.

### D. Hybrid Methods

In classifier fusion, the outputs of individual classifiers are combined by a second classifier according to a pre-defined combination rule. Classifier combination can essentially be implemented at three levels [19]: Fusion at the a) *Feature Extraction level* b) *Confidence or Matching score level* and c) *Decision level*.

The use of classifier fusion has produced many combination techniques over the years. One popular approach has been the idea of bagging [20], which manipulates the training-data with sub-sampling. Another common algorithm, boosting [21], also manipulates the training data, but with emphasis on the training of samples that are difficult to classify

### III. IMPLEMENTATION

In refining the nature of the tasks being evaluated and considering the most influential factors that will adversely affect recognition rates, one establishes five main areas of focus [22]: *Viewing Angle, Illumination, Expression, Occlusion* and *Time Delay*. The AR Database [23] provided the most efficient publicly available source of subject samples, which overlapped most of the desired categories and effectively facilitated testing. The database contains over 3000 images of 135 individuals (76 males and 60 females. The subjects were recorded over two sessions with a two week interval between shoots. During each session 13 images per subject under varying conditions, 1 neutral and 3 per variant of expression, illumination and occlusion (with lighting changes).

Each database image was cropped and resized to dimensions of (125 x 165 x 3) and the pre-processed using the traditional procedure. The preprocessed database formulated a baseline system [5] which was then divided into gallery and probe sets. 110 subjects (55 males and 55 females) were randomly selected, yielding a database of 2860 images. Since it is desirable to have no overlap between the training and testing images, subject images were divided into as follows: the *Training Set* comprised of the neutral expressions of both the 1st and 2nd sessions. This offered all the necessary facial and temporal information and effectively facilitated a realistic investigation of geometrical changes for a small sized training set i.e. expression, illumination and occlusion over time given that few training samples may be available.

The *Test Set* was divided into four categories: Expression, Illumination, Upper Occlusion and Lower Occlusion. Each test set contained 6 images per class/subject (3 per session, 660 images in total per category). In order to achieve a good degree of confidence in the results, each probe set was further subdivided into 10 probe sets comprising 2 random images per subject (220 images per probe set). Although temporal changes are inherently considered in each category due to the session 1 and 2 images, it was also decided to directly compare the effects of time only. A further Time probe set was created, using only the neutral images i.e. images from the 2nd session are tested against the neutral images from the 1st session.

phase, where the comparative assessment will be based on the combinatorial results of three successive tools. Firstly the binomial cumulative probability of correct class assignment will be presented in traditional tabular format. This will be followed by the FERET testing protocol using *Cumulative Match Scores* (CMS) curves also known as rank score [24] and will offer intuitive insight into which algorithm performance throughout the rank spectrum. Finally statistical measures are also applied in the form of McNemar's Hypothesis Protocol [25] that offers the practical insight pertaining to *what point does the difference in performance results actually become significant.*

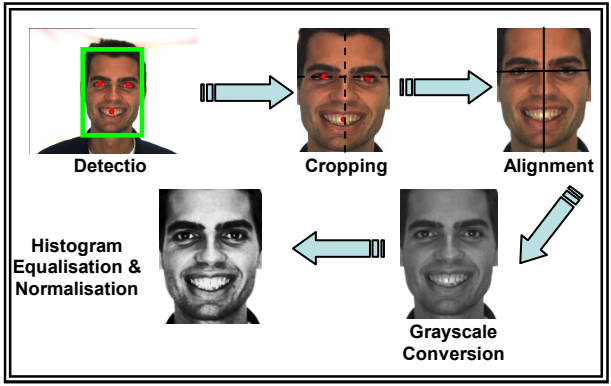

Fig. 1. Depiction of Pre-processing Procedure

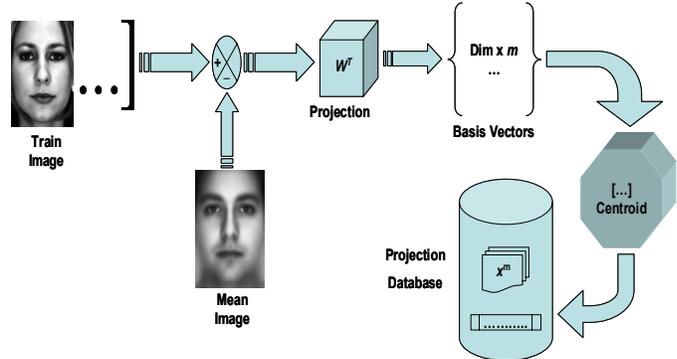

Fig. 2. Generic Training Model

The subspace evaluation is then divided into two aspects, the *training* and *testing*. Training each algorithm followed the procedure in Fig. 2, whereby all the training images are arranged into a column-wise input matrix and sequentially projected into the subspace to yield the projection database. Before any projection or testing could begin however, one needed to obtain a *region of dimensional optimality* for each algorithm i.e. subspace dimensionality selection. Within this paper this region has been defined as a '*Region of constant differentiability, between the similarity measures, which offers comparable accuracies amongst the different algorithms while simultaneously contributing toward effective algorithmic generalisation and computational efficiency*'.

Having no closed form expression to explicitly determine this region, the FERET heuristic was adopted that suggests selecting 40% of the dimensionality would retain approximately 96% of the energy spectrum (signal information) [10]. The upper limit of subspace dimensionality is of course predefined by either the number of training images less one $(M-1)$ in the case of PCA and ICA variants or the number of classes less one $(C-1)$ for LDA algorithms. This was applied to the AR Database where it was found that optimal dimensional retention for PCA and ICA lied between 116 – 199 and between 72 – 74 for LDA.

The testing procedure facilitates the zero-mean Probe Matrices being projected into the same respective subspaces, after which the resulting basis vectors are then compared, using the nearest neighbour approach, to the centroid vectors of class. Each probe image is finally classified with a class label, the results of which are then put forth to the evaluation

Bearing in mind the still, significant image space, $N = 20\,625$, one would like to boost computational efficiency as effectively as possible. Upon initial investigation, in terms of computational resource and time demand, the classical algorithms of PCA and LDA were found to handle this image space very poorly. It was decided, in an effort to significantly reduce the computational load, that the kernel implementations of KPCA and KDA (GDA) will be used to compare both the linear and non-linear projections variants.

Among the non-linear kernel functions, there are many types, the most popular being the Gaussian (RBF) kernel, the Polynomial kernel and the Sigmoidal function. All these functions were investigated and it was found that the Gaussian and second degree Polynomial kernels, empirically achieved the lowest error rates. The final results for the classes of PCA and LDA will hence be assessed amongst three variants: linear, RBF and $2^{nd}$ degree polynomial. Looking at the class of ICA, it has already been mentioned that the Kernel implementation is not feasible for large input spaces; the results will therefore be evaluated demonstrating the algorithms of InfoMax and FastICA in combination with both Architectures I and II.

Upon establishing the 'best' projection-metric combinations for each evaluation category, these approaches will then be combined to realize a fusion strategy that will effectively demonstrate the advantageous results of hybrid formulations. Ross and Jain [19] reported their conclusive findings which indicate that the approach of classification before fusion actually performs poorer than the confidence score level. The matching score level was found to offer the best tradeoffs in terms of information content and ease of fusion [19]. It was therefore selected as the most appropriate level of fusion for this particular application. The similarity measures from the relevant metric of each algorithm will be taken as inputs to the combinational classifier. Normalisation of both the differing metric measures are performed employing the MinMax scheme, resulting in a common range of [0 100].

In combining the different metric measures, the *weighted sum rule* is selected as the fusion rule. Despite its simplicity, the sum rule often outperforms other combination schemes and because of its linear model it is proven to be more tolerant to noise signals, unlike the *product* rule that severely magnifies any noise contributions. The combined matching score will be calculated as follows:

$$S_{comb} = w_1 S'_{FA1} + w_2 S'_{FA2} + w_3 S'_{LDA} \qquad (1)$$

The weights associated with each classifier's matching score are defined by a confidence function to represent the contributional degree of each classifier. This paper proposes two methods that establish these weight values:

*1) Method 1*: This approach is an intuitive generalisation, whereby the weightings are defined as per class performance level in each of the five relevant evaluation categories. If LDA, for example, performs well in the categories of Expression and Time Delay, it will obtain two fifths or 40% of the weighting; similarly if FA1 outperformed the other algorithms in both occlusion categories it will also receive 40% of the weighting; with the remaining 20% going to the algorithm performing best in the category of illumination.

*2) Method 2*: Because recognition accuracy of each component classifier is directly related the confidence one has in its abilities, one can generate a confidence function as a weighting function. Letting $r_i$ be the recognition accuracy of each classifier, the sum of recognition accuracies is given by:

$$r_{sum} = \sum_{i=1}^{q} r_i \qquad (6)$$

where $q$ is the number of classifiers being combined. The associated weightings are then given by:

$$\sum_{i=1}^{q} w_i = \sum_{i=1}^{q} \frac{r_i}{r_{sum}} = 1 \qquad (7)$$

Once the hybrid matching score has been calculated, final classification and evaluation is performed.

## IV. RESULTS AND ANALYSIS

Using the described methodology the relative performance levels of the three most popular classes of appearance based subspace methodologies i.e. PCA, LDA and ICA were investigated. In conjunction with the assessment process one also sort to find the best metric combination that offered the best task specific advantage. One first considered the rank-1 performances of each algorithm and confirmed their performance in accordance with the highest CMS metric curves that offered the highest accuracy levels across the rank spectrum; this either confirmed the rank-1 choices or revealed a metric that offered a better overall performance.

Having found the best projection-metric combination for each algorithm, one then wanted to establish the 'best' algorithm within each subspace class. Although the respective CMS curves offered a meaningful and sometimes very convincing indication of superiority order, it was felt that a deeper analysis would offer greater performance distinction. This was accomplished by employing McNemar's Hypothesis Protocol that aided one in making a decision that is not only intuitively correct but statistically sound as well. This combinatorial approach offered much more significant insight into the relative performance of each variant within the classes and also brought forth the best alternative that one could or should consider for inter-class comparisons.

The inter-class assessments, rank-1 results, were carried out in much the same fashion as the intra-class tests were, the CMS curves were used as the primary tool for obtaining an intuitive indication as to which class performed better and this was confirmed, regionally clarified or nullified by the findings of McNemar's evaluation. The CMS charts, Decision Graphs and Confidence Probabilities were then cohesively used in revealing any significant, task specific, advantage that one class may offer over the next.

In the class of PCA, the following was categorically found:

*1) Expression*: There is no statistical difference between any of the variants, so given the non-rigidity of the facial object, one would still expect a similar performance from the linear and non-linear implementations. The CMS curves do, however, indicate that it is the polynomial variant that offers the slight advantage in recognition rate. Based only on this intuitive deduction, the greater confidence is placed in the polynomial approach. Considering the metric combinations, one found that the best rank-1 results were obtained with the L1 measure and the best overall performance is offered by the L2 measure followed by cosine.

*2) Illumination*: The polynomial approach offered the best CMS accuracy levels, particularly in early ranks, where it was found to be statistically superior in ranks 1 and 2. The L1 measure was found to offer both the best rank-1 and overall performance that was closely followed by the L2 metric.

*3) Lower Occlusion*: These results were by far the lowest achieved and in some ways can even be considered as coincidental. Performance conclusions were reached simply on available evidence but one should leave room for further investigation. From current results, the linear and RBF implementations offered the best performance, with no statistical difference between the two. The CMS curves suggest that early rank advantage is given by the linear variant, while the RBF algorithm claims supremacy after rank-30. The Mahalanobis measure, surprisingly, offered the best metric results for this measure and was closely followed by the City-Block (L1) similarity measure.

*4) Upper Occlusion*: The polynomial variant again found intuitive and statistical superiority across most of the rank spectrum. The best results were found by employing the L1 measure at rank-1 and the Euclidean (L2), followed by the Cosine metrics for overall performance.

*5) Time Delay*: The polynomial algorithm, combined with the L1 metric, once again found early rank supremacy.

On average one could recommend that the best PCA performance levels are offered by the *Polynomial* algorithm in combination with the metrics of L1, followed by L2.

In the class of LDA, the results were as follows:

*1) Expression*: Both non-linear variants of RBF and Polynomial offered equal statistical advantage over the linear approach. The CMS curves were indicative of marginal polynomial superiority and it was therefore selected for inter-class evaluation. The best metric results were reached using the Mahalanobis (Mah) measure, followed by cosine and Euclidean.

*2) Illumination*: The non-linear approaches again show early rank supremacy with the polynomial approach being statistically better than RBF at ranks 1 and 2. The Mahalanobis distance measure was also the best metric measure, followed by L2.

*3) Lower Occlusion*: The linear variant, without question, offered the best statistical and intuitive results across the spectrum. This conclusion however should be considered while also bearing in mind that the performance levels of appearance based methods are extremely sensitive to occlusion, specifically lower facial concealment. The best metric was again found to be the Mahalanobis measure, followed by cosine.

*4) Upper Occlusion*: The best performance was without a doubt offered by the polynomial approach in combination with the Mahalanobis metric.

*5) Time Delay*: There exists no statistical difference between any of the variants when it pertains to temporal face identification. CMS intuition does, however, suggest the marginally better performance being offered by the polynomial algorithm. Again the Mahalanobis measure offers the best results, followed by the cosine metric.

Overall one could suggest that better performance levels were obtained by the non-linear variants, specifically the *Polynomial* approach in combination with the Mahalanobis similarity measure.

In the class of ICA:

*1) Expression*: There was no statistical difference found between any of the ICA variants. *Architecture II*, in both the InfoMax and FastICA implementations, offered the best CMS advantage and *FA2* was selected solely on the fact that FastICA is computationally less costly than the InfoMax alternative. The best combinational metric was provided by the Cosine (*Cos*) measure.

*2) Illumination*: Again there was no statistical difference between any of the algorithms and *FA1* is selected purely on CMS intuition as the alternative that offers the most promising recognition rates. The Cosine metric again, also provides for the best combinatorial results.

*3) Lower Occlusion*: *Architecture I* was found to achieve the best statistical and intuitive early rank results. There was no clear distinction between the FastICA and InfoMax implementation, and again the FastICA variant was selected due to computational advantage. The Cosine metric proved clearly superior once again.

*4) Upper Occlusion*: The FastICA variants were found to perform much better than their InfoMax counterparts, with *Architecture I* reigning supremely and providing distinct rank 2 and 3 superiority over *Architecture II*. The cosine metric by far offers the best results.

*5) Time Delay*: Statistically, there was no significant difference between any of the approaches. *FA1* was however, again selected for inter-class evaluations based only on intuitive CMS performance. Cosine, once again offers the best metric results.

An overview of the ICA class shows that in the categories of expression, illumination and time delay, there is no significant statistical difference between any of the architectures and the choice of employing either the InfoMax or FastICA implementations does not affect the overall performance rankings. In the case of Occlusion, however, *Architecture I* proved the most successful, reiterating the advantage that spatially localised vectors can offer over global (overlapping) feature vectors. In selecting the best metric combination for ICA, the Cosine measure was without a doubt the best distance measure in all categories

In performing the Inter-class assessments the results were as follows:

*1) Expression*: LDA and ICA came out as the top classes, but only being superior to PCA at rank-1; other than that there was no statistical difference between any of the classes. Intuitively one could claim that PCA offers the best CMS result, followed by LDA and lastly ICA, however, this intuition would fall short of true performance conclusions, which in this case is similar for all the classes.

*2) Illumination*: LDA and ICA both claim statistical superiority over PCA for the first 7 ranks; ICA however, outperforms LDA for the first 3 ranks leading one to the conclusion that ICA is the best class to apply for the task of illumination changes.

*3) Lower Occlusion*: While LDA may have outperformed PCA for the first 40 ranks, it was ICA (*Arch I*) that reigned statistically and intuitively supreme throughout the rank spectrum.

*4) Upper Occlusion*: ICA (*Arch I*) again performed the best amongst the three classes. PCA in this category, however, was found to statistically outclass LDA in both early and late rank evaluations.

*5) Time Delay*: Both LDA and ICA outperformed PCA for the first ten ranks. There was no statistical difference between these two classes; however, intuitive analysis would suggest that it is LDA that offers the best performance.

In summing up the class results, while it is true that the specific nature of the task may greatly influence the performance level of any algorithm, on average one could confidently recommend that the class of ICA is perhaps the most flexible and widely adaptable subspace methodology that could be applied, followed by the classes of LDA (non-linear) and PCA (non-linear), respectively.

Two classes of subspace methods stood out, those being the classes of ICA and LDA. The hybrid formulation will thus seek to combine and exploit the powerful data representation of ICA and the unique class discriminality of LDA. Of course the best projection-metric combination from each class is selected so as to develop the most advantageous hybrid performance available. In the LDA class, the polynomial approach in combination with the Mahalanobis similarity measure was selected. In the class of ICA, in the categories of Expression, Illumination and Time delay, it was observed that there were no statistical differences between any of the variants, however FA2 (*Cosine*) did seem intuitively better in the category of Expression and FA1 (*Cosine*) did come out very strong in the categories of Illumination and Time delay; also in the occlusion categories FA1 was clearly the superior algorithm. It was therefore decided to combine both ICA architectures, *Arch I* and *II*, using the computationally more efficient FastICA implementation. By incorporating both architectures, one hoped to harness the advantages offered by both spatially localised and global independent components. Looking at Table I, it is observed that the Hybrid algorithm performs exceptionally well, having the best overall recognition rate performance in four out of the five categories; only in the category of Time Delay did LDA the constituent algorithms. The only advantage that is offered by the Hybrid formulation, over the ICA algorithms, is the marginal superiority in CMS accuracy levels.

*2) Illumination*: Comparing the Hybrid and ICA algorithms, one again finds no statistical difference between the algorithms. When comparing the results to LDA however, one finds that the Hybrid offers a CMS advantage for the first 15 ranks and is statistically superior for the first 4 ranks.

*3) Lower Occlusion*: Again there is no significant difference between the Hybrid and ICA results, but the higher CMS curve and low *p*-values between ranks 1-9 and 20-40 do indicate a higher confidence in the Hybrid performance. When comparing the results to LDA, as one would expect, the Hybrid formulation displays 100% confidence in statistical superiority throughout the rank spectrum.

*4) Upper Occlusion*: The results mimic those found in lower occlusion, whereby no significant difference exists between the Hybrid and ICA algorithms and the Hybrid approach offers complete statistical and intuitive superiority over LDA.

*5) Time Delay*: Statistically there is no difference between any of the algorithms. The Hybrid approach only offers a small early rank accuracy level advantage over the ICA constituents.

In summary, this investigation has proposed an integration

TABLE I
RANK-1 COMPARATIVE HYBRID RESULTS

| | Rank 1 Hybrid Results | | | | | CMS Results | |
| --- | --- | --- | --- | --- | --- | --- | --- |
| | FA1 | FA2 | LDA | Hybrid | | Highest Curve | Same as rank-1 |
| **CATEGORY** | | | | Method 1 | Method 2 | | |
| *Expression* | 79.59% | 80.05% | 83.05% | **83.5%** | 82.36% | Hybrid | Yes |
| *Illumination* | 84.14% | 82.23% | 75.18% | **85.5%** | 84.95% | Hybrid | Yes |
| *Lower Occ* | 27.86% | **28.27%** | 5.54% | 17.5% | 27.77% | Hybrid | No |
| *Upper Occ* | 51.59% | 49.63% | 27.91% | 51,59% | **52.81%** | Hybrid | Yes |
| *Time Delay* | 90.00% | 90.00% | **90.91%** | 90.00% | 90.00% | LDA | Yes |

perform better, but only by a tiny magnitude.

Comparing the Hybrid weighting approaches, although very different, both methods performed very well, with method 1 finding superior claim in the categories of Expression and Illumination and method 2 being the better performer in the Occlusion categories. Both performed equally well in the category of Time Delay. Statistically there is no significant difference between the results of either approach.

Turning to McNemar's analyses, the categorical results were as follows:

*1) Expression*: Statistically there is absolutely no significant difference between the Hybrid results and any of

scheme, which combines the output matching scores of the best categorical subspace methodologies. This was done in an effort to establish a combinatorial algorithm that yields improved recognition accuracies in the realm of face identification. The experimental results, although not vastly superior are nonetheless very encouraging and highlight the fact that combinational strategies can in general lead to more accurate face recognition levels than those achieved by individual classifiers. Although the proposed approach only explores one aspect of hybrid synthesis and the results are not statistically superior to the best categorical constituent algorithms, the framework has been made scalable so that future investigations can easily incorporate and improve other face recognition modules in the quest to realise a truly superior performing hybridised subspace methodology.

## V. CONCLUSION

This research investigation presented a rather rare comparative study of three of the most popular appearance-based face recognition projection classes, PCA, LDA and ICA along with the four most widely accepted similarity measures of City Block (L1), Euclidean (L2), Cosine and the Mahalanobis metrics. Although comparisons between these classes can become fairly complex given the different task natures, the algorithm architectures and the distance metrics that must be taken into account, an important aspect of this study was the completely equal working conditions that were provided in order to facilitate fair and proper comparative levels of evaluation. In doing so, one was able to realise an independent study that evaluated the linear and kernel variants of the respective classes and provided both intuitive and statistical evidence into their comparative standings. This work significantly contributes to prior literary findings, either by verifying previous results, offering further insight into why certain conclusions were made or by providing a better understanding as to why certain claims should be disputed and under which conditions they may hold true. By firstly exploring previous literature with respect to each other and secondly by relating the important findings of this paper to previous works one is able to meet the primary objective in providing an amateur, in the field of face recognition, with a good understanding of publicly available subspace techniques.

## REFERENCES

[1] A. Nes, '*Hybrid Systems for Face Recognition*', Master of Science Graduate Thesis, Faculty of Information Technology, Norwegian University of Science and Technology, 2003.
[2] K. Delac, M. Grgic, S. Grgic, '*Independent Comparative study of PCA, ICA and LDA on the FERET Data set*', Proc. of the 4th International Symposium on Image and Signal Processing and Analysis, pp 289-294, 2005.
[3] M.S. Bartlett, J.R. Movellan, and T.J. Sejnowski, '*Face Recognition by Independent Component Analysis*', IEEE Trans. on Neural Networks, Vol. 13, pp. 1450-1464, 2002.
[4] C. Liu and H. Wechsler, '*Comparative Assessment of Independent Component Analysis (ICA) for Face Recognition*', Second International Conference on Audio and Video-based Biometric Person Authentication, AVBPA'99,Washington D. C., USA, March 22-24, 1999.
[5] K. Baek, B. Draper, J.R. Beveridge, and K. She, '*PCA vs. ICA: A Comparison on the FERET Data Set*', Proc. of the Fourth International Conference on Computer Vision, Pattern Recognition and Image Processing, pp 824-827, 2002.
[6] B. Moghaddam, '*Principal Manifolds and Probabilistic Subspaces for Visual Recognition*', IEEE Trans. on Pattern Analysis and Machine Intelligence, Vol. 24, pp. 780-788, 2002.
[7] J.R. Beveridge, K. She, B. Draper, and G.H. Givens, '*A Nonparametric Statistical Comparison of Principal Component and Linear Discriminant Subspaces for Face Recognition*', Proc. of the IEEE Conference on Computer Vision and Pattern Recognition, pp. 535-542, 2001
[8] A.M. Martinez and A.C Kak, '*PCA versus LDA*', IEEE Transactions on Pattern Analysis and Machine Intelligence, Vol. 23, pp 228-233, 2001.
[9] V. Belhumeur, J. Hespanha, and D. Kriegman. '*Eigenfaces vs. Fisherfaces: Recognition using class specific linear projection*', IEEE Transactions on Pattern Analysis and Machine Intelligence, Vol. 19 pp 711-720, 1997.
[10] B.A. Draper, K.Baek, M.S. Bartlett and J.R Beveridge, '*Recognizing Faces with PCA and ICA*', Computer Vision and Image Understanding, Vol. 91, pp 115-137, 2003.
[11] P.J. Phillips, H. Moon, S.A. Rizvi and P.J. Rauss, '*The FERET evaluation methodology for Face Recognition algorithms*', IEEE Trans. Pattern Analysis and Machine Intelligence, Vol. 22, pp. 1090-1104, 2000
[12] A. K. Jain, X. Lu, and Y. Wang, 'Combining Classifiers for Face Recognition', In Proc. of IEEE International Conference on Multimedia and Expo, pp. 13-16, Baltimore, MD, 2003.
[13] G. Shakhnarovich and B.Moghaddam, '*Face recognition in Subspaces*', Handbook of Face Recognition, Springer-Verlag, 2004.
[14] Y. Adini, Y. Moses, and S. Ullman, '*Face recognition: The problem of compensating for changes in illumination direction*', IEEE Transactions on Pattern Analysis and Machine Intelligence, pp 721-732, 1997.
[15] M. Pavlou, '*Face Kernel Extraction from Local Features*', Doctoral Thesis, Faculty of Engineering and Physical Sciences, University of Manchester, 2005.
[16] M. Yang, '*Kernel Eigenfaces vs. Kernel Fisherfaces: Face Recognition Using Kernel Methods*', In Proc. 5th IEEE International Conference on Automatic and Gesture Recognition, pp 215-220, 2002.
[17] M.A. Aizerman, E.M. Braverman and L.I. Rosonoer, '*Theoretical foundations of the potential function method in pattern recognition learning*', Automation and Remote Control, Vol. 25, pp. 821–837, 1964.
[18] B. Schölkopf and K. R. Müller, '*Nonlinear component analysis as a kernel eigenvalue problem*', Neural Computation, pp 1299-1319, 1998.
[19] A. Ross and A. Jain, '*Information fusion in biometrics*', Pattern Recognition Letters, Vol. 24, pp 2115-2125, 2003.
[20] L. Breiman, '*Bagging Predictors*', Machine Learning, Vol. 26, pp. 123-140, 1996.
[21] Y. Freund, R.E. Shaphire, '*A Decision-theoretic generalization of on-line learning and an application to boosting*', Journal of Computer and System Sciences, Vol. 55, pp. 119-139, 1995.
[22] R. Gross, J. Shi and J. Cohn, '*Quo vadis Face Recognition?*', CMU-RI-TR-01-17, Robotics Institute, Carnegie Mellon University, 2001.
[23] A. R. Martinez and R. Benavente, 'The AR face database. Technical Report', Computer Vision Centre Technical Report, Barcelona, Spain, 1998. http://rvl1.ecn.purdue.edu/ARdatabase.
[24] D.I. Domboulas, '*Infrared Imaging Face Recognition using Nonlinear Kernel-based Classifiers*', Master of Science Graduate Thesis, Naval Postgraduate School, Monterey, California, 2004.
[25] M. Grgic, K. Delac, S. Grgic, '*Face Recognition: Hypothesis Testing across all Ranks*', Technical Report: FER-VCL-TR-2005-02, University of Zagreb, Croatia, 2005.